\pdfoutput=1
\documentclass{article}
\usepackage{arxiv}
\usepackage[numbers,sort]{natbib}
\usepackage{times}
\usepackage{etoolbox}
\usepackage[utf8]{inputenc}
\usepackage[T1]{fontenc}
\usepackage{url}
\usepackage{booktabs}
\usepackage{amsfonts}
\usepackage{nicefrac}
\usepackage{microtype}
\usepackage{mathtools,bm,amssymb,xcolor}
\usepackage{amsthm}

\usepackage{colortbl}
\usepackage{enumitem}
\usepackage{subcaption,placeins}
\usepackage{adjustbox,multirow,dcolumn}
\usepackage{xstring}
\usepackage{graphicx}
\usepackage{natbib}
\usepackage{algorithm}
\usepackage{algpseudocode}
\usepackage{tikz}
\usepackage{pgfplots}
\usepackage{comment}

\usepackage{tabularx}
\definecolor{resultOurs}{gray}{0.94}
\definecolor{resultBand}{gray}{0.975}
\newcommand{\resulttablesetup}{%
    \fontsize{9}{10.5}\selectfont
    \renewcommand{\arraystretch}{1.10}%
    \setlength{\tabcolsep}{2pt}%
    \setlength{\heavyrulewidth}{0.7pt}%
    \setlength{\lightrulewidth}{0.35pt}}

\definecolor{cvprblue}{rgb}{0.21,0.49,0.74}
\usepackage[pagebackref,breaklinks,colorlinks,allcolors=cvprblue]{hyperref}
\makeatletter
\@ifpackageloaded{cvpr}{}{\usepackage[capitalize]{cleveref}}
\makeatother
\newcommand{\cvspace}[1]{}

\newcommand{\authornote}[1]{\textsuperscript{\rm {#1}}}

\usepackage{amsmath,amsfonts,bm}

\def\1{{1}}

\DeclareMathAlphabet{\mathsfit}{\encodingdefault}{\sfdefault}{m}{sl}
\SetMathAlphabet{\mathsfit}{bold}{\encodingdefault}{\sfdefault}{bx}{n}

\makeatletter

\newtoggle{onecolumn}
\AtBeginDocument{%
    \if@twocolumn
        \togglefalse{onecolumn}%
    \else
        \toggletrue{onecolumn}%
    \fi}
\pretocmd{\onecolumn}{\toggletrue{onecolumn}}{}{}
\pretocmd{\twocolumn}{\togglefalse{onecolumn}}{}{}
\newenvironment{columnaligned}
    {\iftoggle{onecolumn}{}{\begin{aligned}}}
    {\iftoggle{onecolumn}{}{\end{aligned}}}
    
\newcommand{\columnalign}{\iftoggle{onecolumn}{}{&}}
\newcommand{\columnmathbreak}{\iftoggle{onecolumn}{\quad}{\\}}
\@ifpackageloaded{cvpr}{}{%
    \setcounter{dbltopnumber}{3}
    
    }

\makeatletter
\@ifpackageloaded{cvpr}{}{%
    \DeclareRobustCommand\onedot{\futurelet\@let@token\@onedot}
    \def\@onedot{\ifx\@let@token.\else.\null\fi}}

\providecommand{\eg}{\emph{e.g\@\onedot}}

\providecommand{\ie}{\emph{i.e\@\onedot}}

\graphicspath{{figures/}}

\usetikzlibrary{external}
\usepgfplotslibrary{colorbrewer}
\pgfplotsset{compat=newest}
\ifdefined\pdfshellescape
  \ifnum\pdfshellescape=1\relax
  \fi
\fi

\newcolumntype{d}{D{.}{.}{3.2}}
\newcolumntype{B}{>{\boldmath\DC@{.}{.}{3.2}}c<{\DC@end}}

\DeclareRobustCommand{\method}{%
    \texorpdfstring{D\textsuperscript{3}-Cache}{D3-Cache}}

\newcommand{\teacache}{TeaCache}
\newcommand{\magcache}{MagCache}
\newcommand{\taylorseer}{TaylorSeer}

\DeclarePairedDelimiter{\parens}{\lparen}{\rparen}

\newcommand{\x}{{x}}

\newcommand{\y}{{y}}
\newcommand{\cond}{{c}}

\newcommand{\modelparams}{\theta}

\newcommand{\fmodel}{f_\modelparams}

\newcommand{\realset}{\mathbb{R}}
\newcommand{\complexset}{\mathbb{C}}

\newcommand{\dmdA}{{A}}
\newcommand{\dmdAred}{\tilde{\dmdA}}

\newcommand{\dmdX}{{X}}
\newcommand{\dmdY}{{Y}}
\newcommand{\dmdU}{{U}}
\newcommand{\dmdV}{{V}}
\newcommand{\dmdW}{{W}}
\newcommand{\dmdPhi}{{\Phi}}
\newcommand{\dmdSigma}{{\Sigma}}
\newcommand{\dmdLambda}{{\Lambda}}
\newcommand{\dmdCoeff}{{b}}
\newcommand{\dmdFreq}{{\omega}}
\newcommand{\ypred}{\hat{\y}}

\newcommand{\maxhist}{H}

\newcommand{\maxskip}{S}
\newcommand{\riskthreshold}{\tau} %
\newcommand{\skipcount}{s}

\newcommand{\normaldist}{\mathcal{N}}

\DeclareMathOperator{\bigO}{\mathcal{O}}

\DeclareMathOperator{\diag}{diag}

\DeclareMathOperator{\realpart}{Re}

\newcommand{\hpad}{\hspace*{\fill}}
\newcommand{\subplotsizeslack}{1}
\NewDocumentCommand{\subplot}{ O{1} O{} m m m }{%
    \hpad%
    \begin{subfigure}{(\textwidth / #1) * \subplotsizeslack}
        \includegraphics[width=\textwidth, #2]{#3}
        \caption{#4}\label{#5}
    \end{subfigure}%
}

\makeatother

\title{%
    \method{}: Accelerating Diffusion Models with Dynamic Mode Decomposition}
\newcommand{\authorblock}[3]{%
    \begin{tabular}[t]{@{}c@{}}
        \textbf{#1}\authornote{#2} \\[-0.1em]
        {\normalfont\footnotesize\ttfamily #3}
    \end{tabular}}
\author{%
    \authorblock{Yanbo Li}{1}{yb.li1@siat.ac.cn} \hspace{1.2em}
    \authorblock{An Xiang}{1}{a.xiang@siat.ac.cn} \hspace{1.2em}
    \authorblock{Xiaoxin Hu}{1}{xx.hu2@siat.ac.cn} \hspace{1.2em}
    \authorblock{Feng Wang}{1}{f.wang3@siat.ac.cn} \\[1.6em]
    \authorblock{Cheng-zhong Xu}{2}{czxu@um.edu.mo} \hspace{1.2em}
    \authorblock{Kejiang Ye}{1}{kj.ye@siat.ac.cn} \hspace{1.2em}
    \authorblock{Xitong Gao}{1,\dag}{xt.gao@siat.ac.cn} \\[1.8em]
    \authornote{1}\,University of Chinese Academy of Sciences \\[0.2em]
    \authornote{2}\,University of Macau \\[0.4em]
    {\small \authornote{\dag}\,Corresponding author}
}
\date{}

\begin{document}

\maketitle

\begin{abstract}
    Diffusion Transformers generate high-quality images and videos,
    but every sampling step requires a full network evaluation.
    Training-free caching reduces this cost by reusing or forecasting
    network outputs across steps, but common approaches use
    per-model calibrated skip criteria or maintain per-layer prediction histories.
    We present \textbf{\method{}}, which applies \underline{D}ynamic Mode
    \underline{D}ecomposition (DMD) to \underline{D}iffusion sampling:
    it fits DMD to a short history of exact velocities and forecasts only
    the velocity that the scheduler consumes.
    A horizon-aware gate decides when to skip:
    whenever the network is evaluated, its output is compared with the
    forecast for that step, and the larger the error, the sooner the network
    is evaluated again, up to a cap on consecutive skipped steps.
    \method{} requires no training or per-model offline calibration curve
    and adds negligible memory.
    It shares one history length and risk tolerance
    across the four evaluated models.
    On four state-of-the-art image and video models, \method{} achieves
    speedups of \(2.6\)--\(3.8\times\) with high fidelity to the base model output.
\end{abstract}

\section{Introduction}\label{sec:intro}
\begin{figure}[!b]
    \centering
    \begin{subfigure}[b]{0.48\textwidth}
        \centering
        \makebox[0.28\linewidth]{\scriptsize \taylorseer{} (\( O=1 \))}%
        \makebox[0.28\linewidth]{\scriptsize \taylorseer{} (\( O=2 \))}%
        \makebox[0.28\linewidth]{\scriptsize \method{}}%
        \makebox[0.15\linewidth]{}%
        \\[-0.2em]
        \includegraphics[width=0.28\linewidth, trim=0 1em 0 0]{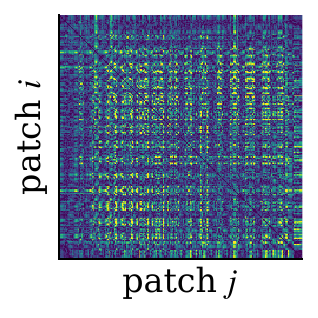}%
        \includegraphics[width=0.28\linewidth, trim=0 1em 0 0]{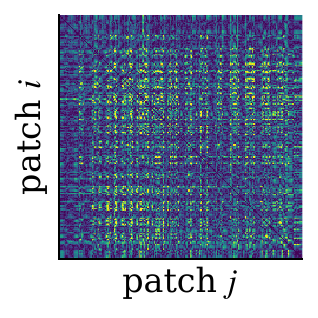}%
        \includegraphics[width=0.28\linewidth, trim=0 1em 0 0]{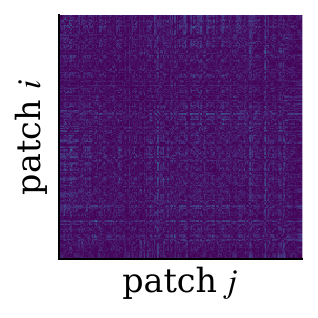}%
        \includegraphics[width=0.15\linewidth, trim=0 0 0 0]{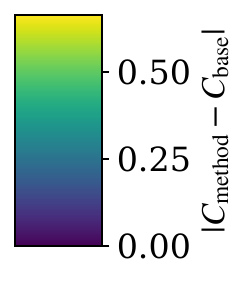}%
        \caption{Spatial residuals (darker $\rightarrow$ lower mismatch)}\label{fig:intro:residuals}
    \end{subfigure}\hfill%
    \begin{subfigure}[b]{0.25\textwidth}
        \centering
        \includegraphics[width=\linewidth, trim=0 1em 0 0]{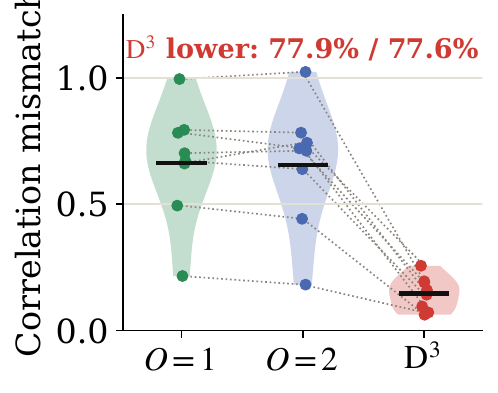}
        \caption{Correlation mismatch}\label{fig:intro:mismatch}
    \end{subfigure}\hfill%
    \begin{subfigure}[b]{0.25\textwidth}
        \centering
        \includegraphics[width=\linewidth, trim=0 1em 0 0]{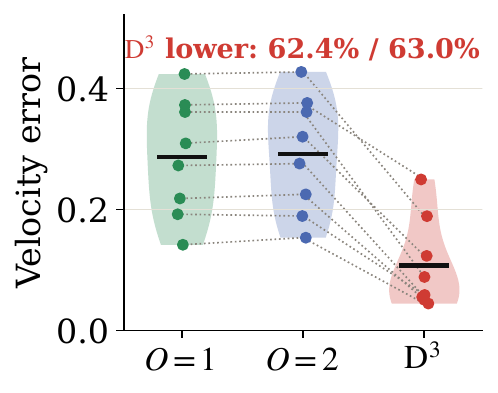}
        \caption{Velocity error}\label{fig:intro:output-error}
    \end{subfigure}
    \caption{
        \textbf{Velocity and patch-correlation diagnostic on FLUX.}
        (a)~Absolute difference between the patch-correlation matrices of
        forecast (\(C_{\mathrm{method}}\)) and exact (\(C_{\mathrm{base}}\)) velocities.
        (b,c)~Per-prompt correlation mismatch and velocity error on eight
        prompts: DMD lowers both relative to first- and second-order
        \taylorseer{} (red: mean reduction relative to \(O=1\) / \(O=2\)).
        Controlled comparisons with shared histories are given in the supplement.
    }\label{fig:intro}
\end{figure}

Diffusion models produce high-quality images and videos through iterative
denoising~\cite{ho20ddpm,song20ddim,rombach21ldm,saharia22imagen,brooks24worldsimulators}.
Diffusion Transformers (DiTs) scale this paradigm to billions of
parameters~\cite{peebles22dit,bao23vit},
but every one of the tens of sampling steps requires a full transformer
evaluation, which dominates generation latency for high-resolution images
and long videos.

Training-free caching exploits the similarity between consecutive steps
to avoid part of this computation~\cite{ma24deepcache,wimbauer24cacheme,selvaraju24fora,zou24toca,zou25duca}.
Reuse methods such as \teacache{} and \magcache{} decide when to skip an
evaluation using change proxies calibrated offline for each
model~\cite{liu24teacache,ma25magcache}.
Forecasting methods such as \taylorseer{} and HiCache extrapolate
intermediate features instead, but store derivative histories in every
layer~\cite{liu25taylorseer,feng25hicache}.
Their memory grows with network depth and expansion order:
on Wan2.1, second-order forecasting raises peak memory from 24.5 to
65.5\,GiB, and on HunyuanVideo it exceeds the GPU memory.
In these methods, skip criteria are not updated from realized forecast errors:
the error of earlier forecasts does not inform later skip decisions.
\begin{figure*}[!t]
    \centering
    \includegraphics[width=0.9\textwidth]{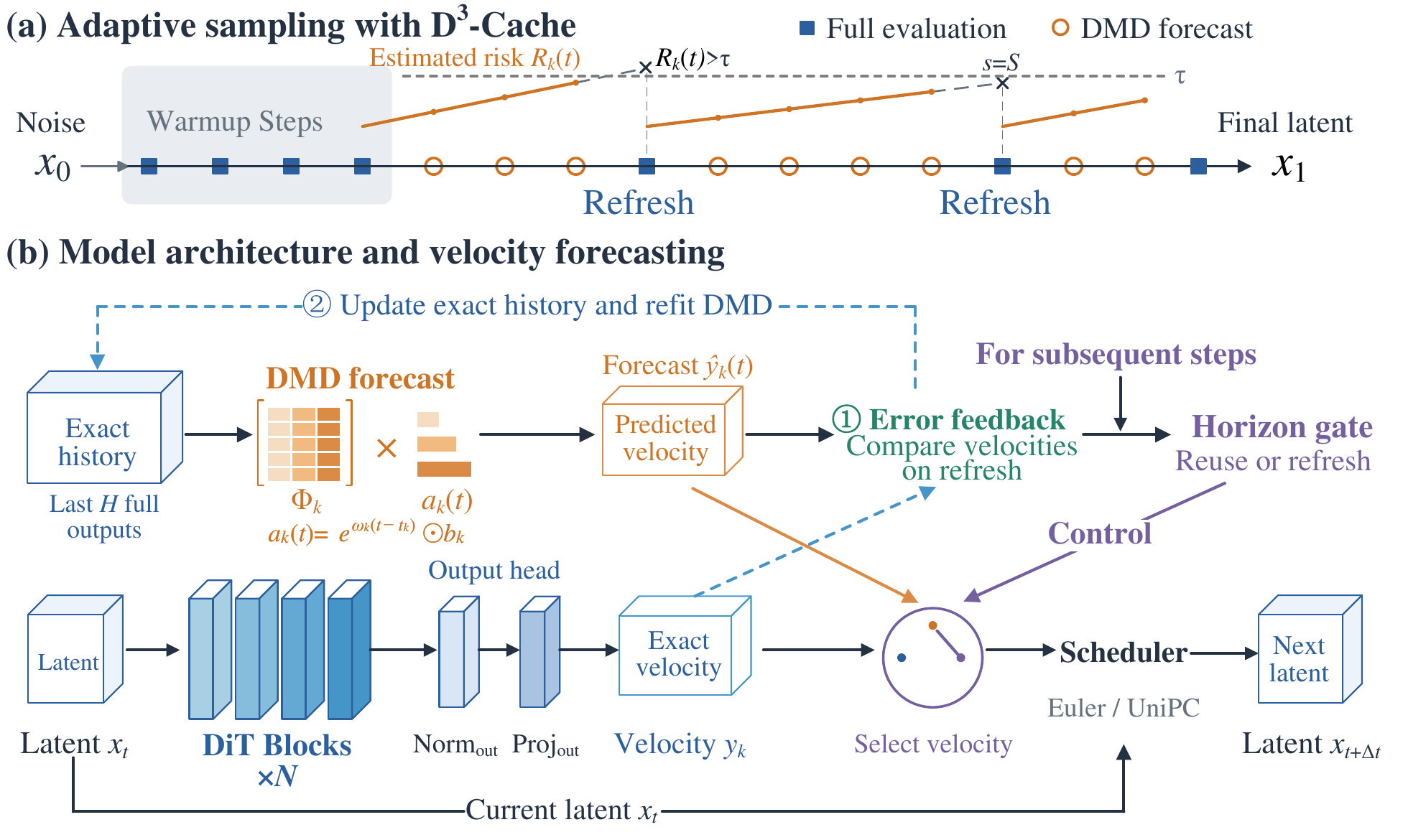}
    \caption{%
        \textbf{Overview of \method{}.}
        (a)~After warmup, each sampling step either runs a full evaluation
        or reuses a DMD forecast of the velocity.
        The estimated risk grows with the horizon since the last full
        evaluation; exceeding \(\riskthreshold\) or reaching the skip cap
        \(\maxskip\) triggers a refresh.
        (b)~At a refresh, the exact velocity first scores the forecast made from
        the preceding history~\textcircled{\scriptsize 1}, which updates the
        risk slope for later steps, and is then pushed into the history used
        for the next DMD fit~\textcircled{\scriptsize 2}.
        For subsequent steps, the gate selects the forecast when
        \(R_k(t)\le\riskthreshold\) and \(\skipcount<\maxskip\); the DiT
        blocks and output head are then skipped entirely.
        Otherwise they are evaluated and the exact velocity is selected.
    }\label{fig:architecture}
\end{figure*}

This memory cost stems from forecasting the features of every layer.
A scheduler step, however, needs only the final output of the network,
the velocity, which has the size of the latent.
Forecasting the velocity directly requires storing only a few latent-sized
snapshots, regardless of network depth.
The velocity trajectory is also highly structured:
patch-level velocity updates are strongly correlated (\cref{fig:intro}).
Dynamic Mode Decomposition (DMD)~\cite{kutz16dmd,dey22koopman} fits
shared temporal modes to the velocity.
On FLUX.1-dev, DMD forecasts of the velocity have lower velocity error and correlation
mismatch than activation forecasts (\cref{fig:intro}); the supplement
separates predictor and state-space effects with shared-history controls.
Finally, every full evaluation reveals the exact velocity at that step,
providing free supervision for the forecaster itself.

Building on these observations, we propose \method{} (\cref{fig:architecture}).
It fits DMD to the most recent exact velocities and supplies the forecast
to the native scheduler in place of a full evaluation.
A horizon-aware gate closes the loop:
at each full evaluation, the forecast made from the preceding history is
scored against the exact velocity, and the resulting horizon-normalized
error rate predicts the risk of extrapolating further.
Forecasts are accepted while this risk stays below a tolerance \(\tau\),
subject to a consecutive-skip cap \(S\).
The gate uses causal error feedback rather than a per-model offline calibration curve.
We select one shared history length \(H\) and tolerance for the four models;
\(S\) selects each operating point.

In summary, our contributions are as follows:
\begin{itemize}
    \item \textbf{Velocity-space DMD forecasting.}
    We forecast the velocity with a low-rank DMD model whose state is
    independent of network depth, characterize its consistency under explicit
    assumptions, and compare predictors at identical full-evaluation positions.
    \item \textbf{Closed-loop, horizon-aware refresh.}
    We score each forecast against the next exact velocity and convert the
    observed error rate into a horizon-dependent risk that governs reuse
    within a skip cap, without a per-model offline calibration curve, and analyze error accumulation
    for Euler sampling conditional on the realized forecast errors.
    \item \textbf{Evaluation on four models.}
    With one shared \(H\) and \(\tau\), \method{} accelerates four image and
    video diffusion models by \(2.6\)--\(3.8\times\) and stays closer to the
    full-step output than the compared caching methods at comparable speed,
    with negligible memory overhead.
\end{itemize}

\section{Related work}\label{sec:related}

\paragraph{Diffusion models and sampling acceleration.}
Diffusion models generate samples through iterative denoising or score-based
dynamics~\cite{sohldickstein15thermo,ho20ddpm,song19score,dhariwal21beatgans}.
Latent representations and transformer backbones support scalable image and
video synthesis~\cite{rombach21ldm,peebles22dit,ma25latte,zhao25pab}.
DDIM and numerical solvers reduce sampling evaluations~\cite{song20ddim,karras22edm,lu25dpmsolver};
distillation and consistency training learn short sampling
trajectories~\cite{salimans22distillation,song23consistency,luo23lcm}.
\method{} is complementary: it keeps the native sampling schedule and
solver, and replaces selected network evaluations with forecasts,
without additional training.

\paragraph{Feature reuse and forecasting.}
DeepCache, Cache Me if You Can, Faster Diffusion, FasterCache and T-GATE reuse
block, branch or attention computations across sampling
steps~\cite{ma24deepcache,wimbauer24cacheme,li24fasterdiffusion,lv24fastercache,zhang24tgate}.
FORA, ToCa, DuCa, \(\Delta\)-DiT and AdaCache extend caching to
DiTs~\cite{selvaraju24fora,zou24toca,zou25duca,chen24deltadit,kahatapitiya24adacache}.
TeaCache maps changes in timestep-embedding-modulated inputs to an estimated
output change using a calibrated polynomial~\cite{liu24teacache}.
MagCache models accumulated reuse error using calibrated successive-step
residual-magnitude ratios, a threshold and a skip cap~\cite{ma25magcache}.
Both criteria are fitted offline and are not updated by the forecast error
of the current sample.
TaylorSeer and HiCache forecast internal activations using Taylor and
scaled-Hermite expansions~\cite{liu25taylorseer,feng25hicache};
their per-layer histories grow with network depth and expansion order.
\method{} instead forecasts only the scheduler input and adapts its refresh
decisions online from observed forecast errors.

\paragraph{Velocity estimation.}
Velocity decomposition and estimation (VDE) splits the velocity into components parallel and orthogonal to the
current latent, linearly extrapolates scalar coefficients, and reuses a
recent orthogonal direction~\cite{tan26vde}.
\method{} instead fits a low-rank linear operator to the full velocity
history, so its forecasts evolve all coordinates jointly through shared
temporal modes, and couples this predictor with closed-loop,
horizon-aware refresh control.

\paragraph{Koopman operators in deep learning.}
Koopman operator theory, which underlies DMD~\cite{tu14dmd,kutz16dmd},
has been used to forecast network weights during training~\cite{dogra20koopman}
and to learn linearized generative dynamics for few-step
sampling~\cite{bai25koopmandiffusion,turan25koopmanflow}.
These methods train dedicated models, whereas \method{} fits DMD online to
the sampling trajectory of a pretrained DiT.

\section{Method}\label{sec:method}

\method{} accelerates sampling by forecasting the velocity supplied to the
scheduler between full transformer evaluations.
A DMD predictor models the local temporal evolution of this velocity,
and a closed-loop, horizon-aware gate decides when to refresh the forecast
with a new full evaluation.

\subsection{Preliminaries}\label{sub:method:preliminaries}

\paragraph{Rectified flow models.}
Rectified flow transports a noise distribution \(\pi_0\)
to a data distribution \(\pi_1\) through a learned velocity field~\cite{liu22rectifiedflow}.
For conditional generation, let \(\x_0\sim\pi_0=\normaldist(0,I)\)
and \(\x_1\sim\pi_1(\cdot\mid c)\) be independent noise and data latents
in \(\realset^D\), where \(c\) is the condition.
The velocity field \(v_\modelparams\) is trained on the linear interpolation
\(\x_t=(1-t)\x_0+t\x_1\), \(t\in[0,1]\), to regress \(\x_1-\x_0\).
At inference, the latent follows
\(\mathrm{d}\x_t/\mathrm{d}t=v_\modelparams(\x_t,t,c)\) from \(\x_0\sim\pi_0\);
this trajectory is generally curved, so its velocity varies during sampling.
At sampling time \(t\), an Euler step of size \(\Delta t>0\) takes the form
\begin{equation}
    \x_{t+\Delta t}=\x_t+\Delta t\,v_\modelparams(\x_t,t,c).
    \label{eq:prelim:flow-step}
\end{equation}
The step size follows the sampling schedule.
This separates velocity evaluation from latent integration:
\method{} approximates selected velocity evaluations
while retaining the model's sampling steps and solver.

\paragraph{Dynamic mode decomposition.}
DMD approximates snapshot dynamics with a linear operator~\cite{tu14dmd}.
Given \(m+1\) observations \(z_0,\ldots,z_m\in\realset^n\),
we form \(m\) snapshot pairs in the matrices
\begin{equation}
    \begin{columnaligned}
        \dmdX\columnalign=[z_0,\ldots,z_{m-1}]\in\realset^{n\times m},
        \columnmathbreak
        \dmdY\columnalign=[z_1,\ldots,z_m]\in\realset^{n\times m}.
    \end{columnaligned}
    \label{eq:xy}
\end{equation}
The minimum-norm least-squares map is
\(\dmdA=\dmdY\dmdX^{\dagger}\), so that \(\dmdY\approx\dmdA\dmdX\),
where \(\dagger\) denotes the Moore--Penrose pseudoinverse.
Using a rank-\(r\) singular value decomposition (SVD)
\(\dmdX\approx\dmdU_r\dmdSigma_r\dmdV_r^\top\),
with \(r\le\min(n,m)\), DMD computes a reduced operator and its modes:
\begin{equation}
    \begin{aligned}
        \dmdAred&=\dmdU_r^\top\dmdY\dmdV_r\dmdSigma_r^{-1},
        &\dmdAred\dmdW&=\dmdW\dmdLambda,\\
        \dmdPhi&=\dmdY\dmdV_r\dmdSigma_r^{-1}\dmdW.
    \end{aligned}
    \label{eq:dmd-core}
\end{equation}
Here, \(\dmdW\in\complexset^{r\times r}\) contains the reduced eigenvectors,
\(\dmdLambda\in\complexset^{r\times r}\) is diagonal with their eigenvalues,
and the columns of \(\dmdPhi\in\complexset^{n\times r}\) are the DMD modes.
Each mode describes a pattern across snapshot coordinates;
its eigenvalue governs the pattern's temporal evolution.
This representation allows the output coordinates to evolve jointly
through a small set of temporal coefficients.

\subsection{DMD for velocity forecasting}\label{sub:method:dmd-cache}

We apply DMD to the velocity state supplied to the native scheduler,
as illustrated in \cref{fig:architecture}.
At sampling time \(t\), a complete velocity evaluation returns
\begin{equation}
    \y=\fmodel(\x_t,t_{\mathrm{model}},\cond)\in\realset^D,
    \label{eq:prelim:diffusion}
\end{equation}
where \(\x_t\) is the current latent,
\(t_{\mathrm{model}}\) is the model's input timestep corresponding to \(t\),
and \(\cond\) is the condition.
Here \(\fmodel\) denotes the velocity field \(v_\modelparams\) together with the model's
output conversion and guidance, so \(\y\) is exactly the state consumed by
the native scheduler.
For models with classifier-free guidance (CFG)~\cite{ho22cfg}, \(\y\) is the guided velocity;
for models with embedded guidance, it is the model output.
A forecast therefore replaces the entire guided evaluation
and enters the scheduler through the same interface.
The velocity tensor is flattened for fitting,
so the cache stores at most \(\maxhist\) tensors of the latent size \(D\),
independent of network depth.
Scheduler-specific time and sign conventions are given in the supplement.

\paragraph{Local velocity dynamics.}
We model local velocity evolution in the forward time \(t=1-\sigma\),
where \(\sigma\) is the scheduler's noise level, so that horizons are the
actual decreases in \(\sigma\) rather than differences of timestep indices.
Let \(k\) count full evaluations and \(t_k\) denote the time of the \(k\)-th one.
These times form a subsequence of the sampling schedule;
\(k\) does not increase at predicted steps.
We retain at most \(H\ge2\) exact velocities.
After at least two full evaluations, the history consists of
\begin{equation}
    (t_j,\y_j),\quad j=k-m,\ldots,k,\quad m=\min(k,H)-1.
    \label{eq:history}
\end{equation}
These \(m+1\) velocities form \(m\) snapshot pairs
in \cref{eq:xy,eq:dmd-core}, with \(n=D\) and \(r\) the numerical rank
of \(\dmdX\), so \(r=m\) when \(\dmdX\) has full column rank.
The fitted factors \(\dmdPhi_k,\dmdLambda_k\)
represent the local evolution of the output trajectory.

\paragraph{Forecasting between full evaluations.}
Refresh times can be irregular, so we approximate their spacing
by the mean interval within the current history:
\begin{equation}
    \bar{\Delta t}_k=\frac{t_k-t_{k-m}}{m},\qquad
    \dmdFreq_k=\frac{\log(\diag(\dmdLambda_k))}{\bar{\Delta t}_k}.
    \label{eq:omega}
\end{equation}
The frequencies \(\dmdFreq_k\) and amplitudes
\(\dmdCoeff_k=\dmdPhi_k^\dagger\y_k\) give the forecast at \(t>t_k\):
\begin{equation}
    \ypred_k(t)=\realpart\parens*{
        \dmdPhi_k\bigl(\exp(\dmdFreq_k(t-t_k))\odot\dmdCoeff_k\bigr)}.
    \label{eq:predict}
\end{equation}
Here, \(\diag\) extracts the diagonal, \(\log\) (principal branch) and \(\exp\) act element-wise,
\(\realpart\) takes the real part, and \(\odot\) denotes element-wise multiplication.
The modes remain fixed between full evaluations,
while their shared coefficients evolve over the elapsed span \(t-t_k\).
A new full evaluation adds an exact velocity to the history
and updates the fitted dynamics; predicted outputs do not enter this history.
The history length sets the predictor capacity:
with \(\maxhist=2\), the fitted operator has rank one and the forecast
rescales \(\y_k\) along its own direction;
additional independent snapshots can add modes, allowing the forecast to change
direction within the span of recent velocities (derivation in the supplement).
Fitting requires only a thin SVD of a \(D\times m\) matrix,
whose \(\bigO(Dm^2)\) cost accounts for less than 1\% of end-to-end
latency on all four models (supplement).

\subsection{Horizon-aware caching policy}\label{sub:method:adaptive_caching}

The gate determines how far to extrapolate from the latest full evaluation
using previously measured forecast errors and the requested prediction span.
At a full evaluation \(k\) with a preceding forecast,
we compare the exact velocity \(\y_k\)
with the forecast \(\ypred_{k-1}(t_k)\) made from the preceding history:
\begin{equation}
    e_k=\frac{\|\ypred_{k-1}(t_k)-\y_k\|_2}
                   {\|\y_k\|_2},\qquad
    q_k=\frac{e_k}{t_k-t_{k-1}}.
    \label{eq:gate-observation}
\end{equation}
The rate \(q_k\) normalizes the relative error by its forecast horizon.
Let \(\mathcal I_k\) index the three most recent full evaluations
with valid error rates \(q_j\).
Let \(\tilde q_k\) denote the median of these recent rates
\(\{q_j\}_{j\in\mathcal I_k}\).
The risk at a subsequent step \(t>t_k\) is
\begin{equation}
    R_k(t)=\tilde q_k(t-t_k).
    \label{eq:gate-risk}
\end{equation}
The risk grows with the extrapolation span at the observed error rate,
so the gate adapts the forecast interval to both recent prediction
accuracy and the distance from the latest exact observation.
Because \(R_k\) is computed solely from errors already observed at full
evaluations, the gate is causal and needs no per-model offline calibration curve.
It is an empirical risk estimate, not an upper bound on future error.

We use full evaluations during the initial warmup steps
to initialize the velocity history and collect error observations,
avoiding extrapolation from limited history at the start of sampling.
After initialization, a forecast is used when
\begin{equation}
    R_k(t)\le\riskthreshold,\qquad \skipcount<\maxskip,
    \label{eq:gating}
\end{equation}
where \(\riskthreshold\) is the risk tolerance,
\(\skipcount\) counts consecutive predicted steps,
and \(\maxskip\) limits their number.
Otherwise, the transformer is evaluated and the predictor is refreshed.
The final step also uses a full evaluation.
Within the cap, the error measured at each refresh therefore governs the
decisions until the next refresh.

\paragraph{Sampling procedure.}
\Cref{fig:architecture} summarizes the sampling loop, and the supplement
gives its pseudocode.
At a refresh, the exact velocity first scores the preceding forecast and then
enters the history; since guidance is part of the forecast state,
an accepted forecast skips the entire guided evaluation.

\paragraph{Theoretical properties.}
Under full-rank, nonzero-spectrum and diagonalizability assumptions,
the supplement proves interpolation at the latest exact velocity and a
least-squares recurrence one mean interval ahead.
Exact exponential recovery additionally requires uniform observation spacing
and an unaliased spectrum.
For Euler sampling, a perturbation bound relates latent deviation to the
realized errors at accepted steps; the gate does not certify these unobserved
errors or a bound for the native multistep solvers.

\section{Experimental results}\label{sec:results}

\subsection{Evaluation protocol}\label{sec:results:setup}
\paragraph{Models and benchmarks.}
We evaluate two image models, FLUX.1-dev~\cite{labs24flux} and
HiDream-I1-full~\cite{cai25hidream}, and two video models,
Wan2.1-T2V-1.3B~\cite{wan25wan} and HunyuanVideo~\cite{kong24hunyuanvideo},
on RTX PRO 6000D GPUs with 84\,GB of memory.
All cache-based methods keep the native 50-step schedule and solver.
FLUX and HiDream generate \(1024\times1024\) images with guidance 3.5/Euler
and guidance 5/UniPC~\cite{zhao23unipc}, respectively, on two prompt sets: all 200 DrawBench
prompts~\cite{saharia22imagen} and 100 captions sampled from COCO val2017~\cite{lin14coco}.
Settings are selected on DrawBench and transferred to COCO unchanged.
To cover both prompt sets, the main tables report HiDream on DrawBench
and FLUX on COCO; the supplement reports the other two combinations.
Wan and HunyuanVideo generate 65-frame clips at 16\,fps
from 128 prompts drawn from the official VBench lists~\cite{huang24vbench}:
Wan at \(832\times480\) with CFG 5 and UniPC (shift 5),
and HunyuanVideo at \(640\times480\) with Euler (shift 7)
and embedded guidance 6.
All methods share the same prompts and seeds within each model.
Warmup is four full steps for images and ten for videos respectively.

\paragraph{Baselines.}
We compare with TeaCache~\cite{liu24teacache}, MagCache~\cite{ma25magcache},
TaylorSeer~\cite{liu25taylorseer}, HiCache~\cite{feng25hicache}
and VDE~\cite{tan26vde}, using each paper's notation
(\(\delta\): threshold, \(K\): reuse cap, \(\mathcal N\)/\(n\): interval, \(O\): order).
Baseline settings are chosen from their sweeps near the speed of each
\method{} operating point, except TeaCache on HunyuanVideo, which uses its
official threshold; discrete speed gaps remain, and the supplement details implementations.
For TaylorSeer and HiCache, we report the second-order setting for its
higher fidelity; on HunyuanVideo it exceeds the GPU memory, so we also
report the first-order setting there.

\paragraph{Our configurations.}
All four models share \(H=3\) and \(\tau=0.56\); the skip cap \(S\) sets
the operating point: \(S=7\) for FLUX, \(4\) for HiDream, and \(5\) for
Wan and HunyuanVideo.
The supplement describes how these values were chosen.

\paragraph{Metrics.}
SSIM~\cite{wang04ssim}, PSNR and LPIPS~\cite{zhang18lpips} measure fidelity to the 50-step output with the
same prompt and seed, averaged over eight uniformly spaced frames for videos.
For images, we also report rFID, the FID~\cite{heusel17fid} computed against these
50-step outputs rather than real images.
Within each video table, all completed methods are evaluated on the same
128 prompts.
For videos, we also report the official VBench total~\cite{huang24vbench}
over all 128 prompts.
For images, we also report ImageReward (IR)~\cite{xu23imagereward} and CLIP score~\cite{radford21clip}, which need no reference.
Latency covers text encoding, denoising and decoding;
speedup is relative to the 50-step model of the same implementation,
and VRAM is the peak allocated GPU memory during generation.
We test differences with a paired bootstrap over prompts (10{,}000 resamples)
and call them significant when the 95\% confidence interval excludes zero.

\subsection{Image generation}\label{sec:results:text_to_image_generation}
\begin{table*}[t]
\centering
\caption{\textbf{Quantitative comparison in image generation on HiDream-I1-full (200 DrawBench prompts, 50 steps, 1024\,\(\boldsymbol{\times}\)\,1024).} Fidelity is measured against the 50-step output; IR: ImageReward. Bold marks the best cache-based method.}
\label{tab:hidream-main}
\resulttablesetup
\begin{tabularx}{\textwidth}{@{}lccc*{6}{>{\centering\arraybackslash}X}@{}}
\toprule
\multirow{2}{*}{\textbf{Method}} & \multicolumn{3}{c}{\textbf{Efficiency}} & \multicolumn{4}{c}{\textbf{Fidelity}} & \multicolumn{2}{c}{\textbf{Perceptual}} \\
\cmidrule(lr){2-4}\cmidrule(lr){5-8}\cmidrule(lr){9-10}
 & \textbf{Latency (s)\(\downarrow\)} & \textbf{Speedup\(\uparrow\)} & \textbf{VRAM (GiB)\(\downarrow\)} & \textbf{SSIM\(\uparrow\)} & \textbf{PSNR\(\uparrow\)} & \textbf{LPIPS\(\downarrow\)} & \textbf{rFID\(\downarrow\)} & \textbf{IR\(\uparrow\)} & \textbf{CLIP\(\uparrow\)} \\
\midrule
Vanilla (50 steps) & 63.15 & 1.000\(\times\) & 59.89 & -- & -- & -- & -- & 1.271 & 28.84 \\
Vanilla (25 steps) & 31.75 & 1.989\(\times\) & 59.89 & 0.6201 & 14.22 & 0.3991 & 67.02 & 1.247 & 28.52 \\
\midrule
TeaCache (\(\delta=0.36\)) & 26.96 & 2.342\(\times\) & 59.93 & 0.7360 & 18.39 & 0.2319 & 43.87 & 1.231 & 28.74 \\
MagCache (\(\delta=0.08,K=4\)) & 26.95 & 2.343\(\times\) & 59.93 & 0.7723 & 19.75 & 0.1889 & 36.39 & 1.241 & 28.75 \\
TaylorSeer (\(\mathcal{N}=3,O=2\)) & 26.52 & 2.381\(\times\) & 72.19 & 0.6510 & 15.80 & 0.3477 & 60.43 & 1.198 & 28.78 \\
HiCache (\(\mathcal{N}=3,O=2\)) & 26.25 & 2.406\(\times\) & 72.19 & 0.6736 & 16.54 & 0.3157 & 55.43 & 1.203 & 28.65 \\
VDE (\(n=3\)) & 25.49 & 2.478\(\times\) & 60.15 & 0.7172 & 17.75 & 0.2557 & 46.80 & 1.231 & 28.77 \\
\rowcolor{resultOurs}
\textbf{\method{} (Ours)} & \textbf{24.34} & \textbf{2.594\(\times\)} & \textbf{59.89} & \textbf{0.8551} & \textbf{24.51} & \textbf{0.1112} & \textbf{22.63} & \textbf{1.248} & \textbf{28.86} \\
\bottomrule
\end{tabularx}
\end{table*}

\begin{table*}[t]
\centering
\caption{\textbf{Quantitative comparison in image generation on FLUX.1-dev (100 COCO captions, 50 steps, 1024\,\(\boldsymbol{\times}\)\,1024).} \(\dagger\): run in its native pipeline; latency normalized to the common 50-step baseline.}
\label{tab:flux-coco-main}
\resulttablesetup
\begin{tabularx}{\textwidth}{@{}lccc*{6}{>{\centering\arraybackslash}X}@{}}
\toprule
\multirow{2}{*}{\textbf{Method}} & \multicolumn{3}{c}{\textbf{Efficiency}} & \multicolumn{4}{c}{\textbf{Fidelity}} & \multicolumn{2}{c}{\textbf{Perceptual}} \\
\cmidrule(lr){2-4}\cmidrule(lr){5-8}\cmidrule(lr){9-10}
 & \textbf{Latency (s)\(\downarrow\)} & \textbf{Speedup\(\uparrow\)} & \textbf{VRAM (GiB)\(\downarrow\)} & \textbf{SSIM\(\uparrow\)} & \textbf{PSNR\(\uparrow\)} & \textbf{LPIPS\(\downarrow\)} & \textbf{rFID\(\downarrow\)} & \textbf{IR\(\uparrow\)} & \textbf{CLIP\(\uparrow\)} \\
\midrule
Vanilla (50 steps) & 31.12 & 1.000\(\times\) & 33.81 & -- & -- & -- & -- & 1.077 & 26.25 \\
Vanilla (25 steps) & 15.72 & 1.980\(\times\) & 33.81 & 0.7581 & 18.37 & 0.2831 & 63.11 & 1.043 & 26.34 \\
\midrule
TeaCache (\(\delta=0.95\)) & 8.63 & 3.604\(\times\) & 33.86 & 0.7484 & 19.21 & 0.2934 & 66.86 & 1.023 & 26.15 \\
MagCache (\(\delta=0.16,K=6\)) & 8.34 & 3.733\(\times\) & 33.84 & 0.8019 & 21.29 & 0.2190 & 50.71 & 1.028 & 26.13 \\
TaylorSeer (\(\mathcal{N}=6,O=2\)) & 8.29 & 3.753\(\times\) & 39.82 & 0.6977 & 17.77 & 0.3317 & 66.18 & 1.053 & 26.24 \\
HiCache (\(\mathcal{N}=6,O=2\)) & 8.64\(^{\dagger}\) & 3.601\(\times\) & 42.14 & 0.7190 & 18.60 & 0.3009 & 63.61 & 1.049 & \textbf{26.39} \\
VDE (\(n=5\)) & 8.90 & 3.494\(\times\) & \textbf{33.81} & 0.7450 & 19.39 & 0.2736 & 63.48 & 1.044 & 26.15 \\
\rowcolor{resultOurs}
\textbf{\method{} (Ours)} & \textbf{8.23} & \textbf{3.780\(\times\)} & 33.82 & \textbf{0.8052} & \textbf{21.62} & \textbf{0.2177} & \textbf{50.53} & \textbf{1.054} & 26.22 \\
\bottomrule
\end{tabularx}
\end{table*}

\Cref{tab:hidream-main,tab:flux-coco-main} compare the selected operating points on DrawBench (HiDream) and COCO (FLUX).

\paragraph{HiDream-I1.}
\method{} delivers the highest fidelity on all four metrics at the highest
speedup, \(2.59\times\) (\cref{tab:hidream-main}).
Relative to MagCache (\(2.34\times\)), the strongest baseline, SSIM rises
from 0.772 to 0.855, PSNR from 19.75 to 24.51\,dB, LPIPS falls by 41\% from
0.189 to 0.111, and rFID from 36.4 to 22.6.
The margin widens against activation forecasters: TaylorSeer and HiCache
reach LPIPS 0.348 and 0.316 while using 12.3\,GiB more memory.
VDE, the next fastest method (\(2.48\times\)), has 2.3 times the LPIPS.
The gap persists across operating points (supplement).

\paragraph{FLUX.1-dev.}
On COCO captions, at a matched speed (\(3.78\times\) versus
\(3.49\)--\(3.75\times\)), \method{} attains the best average scores on all four fidelity metrics
(\cref{tab:flux-coco-main}).
Against TaylorSeer (\(3.75\times\)), it gains almost 4\,dB PSNR, lowers LPIPS
from 0.332 to 0.218 and rFID from 66.2 to 50.5, and uses 6.0\,GiB less memory.
Against MagCache (\(3.73\times\)), the strongest baseline, the margins are
small: SSIM 0.802 to 0.805, PSNR 21.29 to 21.62\,dB and LPIPS 0.2190 to
0.2177, and none of these differences is statistically significant.
\method{} uses the \(H\), \(\tau\) and \(S\) selected on DrawBench, and the
same \(H\) and \(\tau\) as on HiDream.

\begin{figure*}[!t]
\centering
\includegraphics[width=\textwidth]{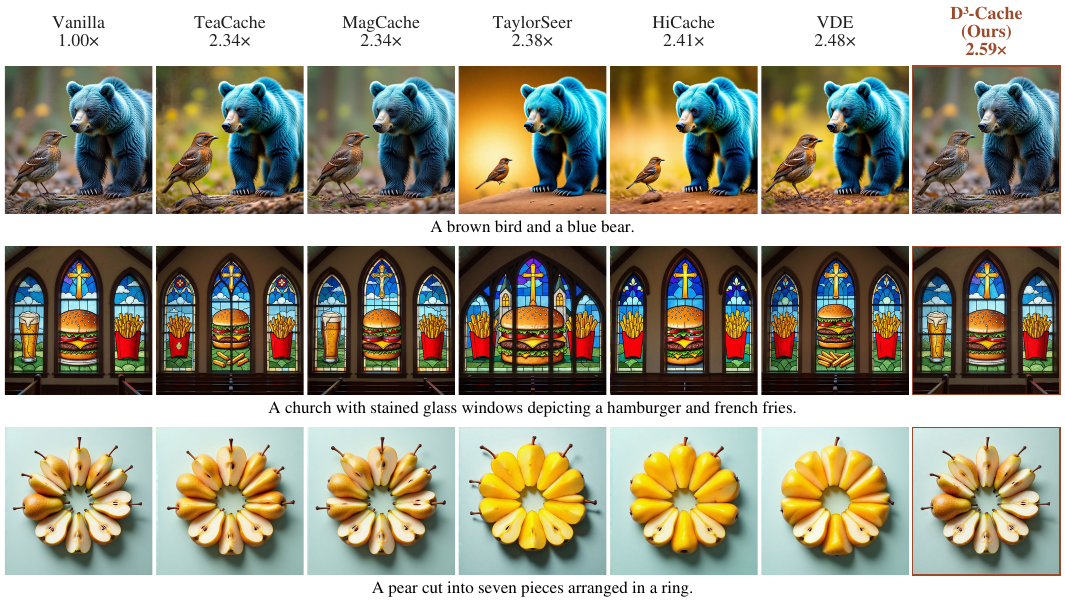}
\caption{\textbf{Qualitative comparison with previous acceleration methods on HiDream-I1 (DrawBench prompts).}
Compared methods change the background, the window contents or the pear texture, whereas \method{} stays closest to the 50-step output.
Examples are taken from the prompts with the largest LPIPS advantage of \method{}; speedups are those of \cref{tab:hidream-main}.}
\label{fig:revision-image-qualitative}
\end{figure*}

\subsection{Video generation}\label{sec:results:text_to_video_generation}
\begin{table*}[t]
\centering
\caption{\textbf{Quantitative comparison in video generation on HunyuanVideo (128 VBench prompts, 50 steps, 640\,\(\boldsymbol{\times}\)\,480, 65 frames).} Second-order TaylorSeer and HiCache run out of GPU memory on this model, so their first-order settings are reported.}
\label{tab:hunyuan-main}
\resulttablesetup
\begin{tabularx}{\textwidth}{@{}lccc*{4}{>{\centering\arraybackslash}X}@{}}
\toprule
\multirow{2}{*}{\textbf{Method}} & \multicolumn{3}{c}{\textbf{Efficiency}} & \multicolumn{3}{c}{\textbf{Fidelity}} & \textbf{Quality} \\
\cmidrule(lr){2-4}\cmidrule(lr){5-7}\cmidrule(lr){8-8}
 & \textbf{Latency (s)\(\downarrow\)} & \textbf{Speedup\(\uparrow\)} & \textbf{VRAM (GiB)\(\downarrow\)} & \textbf{SSIM\(\uparrow\)} & \textbf{PSNR\(\uparrow\)} & \textbf{LPIPS\(\downarrow\)} & \textbf{VBench\(\uparrow\)} \\
\midrule
Vanilla (50 steps) & 275.56 & 1.000\(\times\) & 60.17 & -- & -- & -- & 82.64 \\
Vanilla (25 steps) & 141.56 & 1.947\(\times\) & 60.17 & 0.7946 & 22.51 & 0.2050 & 81.60 \\
\midrule
TeaCache (\(\delta=0.15\)) & 127.69 & 2.158\(\times\) & 60.52 & 0.8232 & 24.54 & 0.1602 & 81.64 \\
MagCache (\(\delta=0.30,K=6\)) & 104.74 & 2.631\(\times\) & 60.28 & 0.9007 & 29.41 & 0.0871 & 81.84 \\
TaylorSeer (\(\mathcal{N}=6,O=1\)) & 101.85 & 2.706\(\times\) & 70.12 & 0.8184 & 26.39 & 0.1091 & \textbf{82.52} \\
HiCache (\(\mathcal{N}=6,O=1\)) & \textbf{100.55} & \textbf{2.740\(\times\)} & 70.12 & 0.8064 & 26.61 & 0.1202 & 82.03 \\
VDE (\(n=5\)) & 104.94 & 2.626\(\times\) & \textbf{60.17} & 0.8510 & 26.24 & 0.1011 & 80.98 \\
\rowcolor{resultOurs}
\textbf{\method{} (Ours)} & 101.55 & 2.713\(\times\) & \textbf{60.17} & \textbf{0.9135} & \textbf{29.74} & \textbf{0.0693} & 82.06 \\
\bottomrule
\end{tabularx}
\end{table*}

\begin{table*}[t]
\centering
\caption{\textbf{Quantitative comparison in video generation on Wan2.1-T2V-1.3B (128 VBench prompts, 50 steps, 832\,\(\boldsymbol{\times}\)\,480, 65 frames).}}
\label{tab:wan-main}
\resulttablesetup
\begin{tabularx}{\textwidth}{@{}lccc*{4}{>{\centering\arraybackslash}X}@{}}
\toprule
\multirow{2}{*}{\textbf{Method}} & \multicolumn{3}{c}{\textbf{Efficiency}} & \multicolumn{3}{c}{\textbf{Fidelity}} & \textbf{Quality} \\
\cmidrule(lr){2-4}\cmidrule(lr){5-7}\cmidrule(lr){8-8}
 & \textbf{Latency (s)\(\downarrow\)} & \textbf{Speedup\(\uparrow\)} & \textbf{VRAM (GiB)\(\downarrow\)} & \textbf{SSIM\(\uparrow\)} & \textbf{PSNR\(\uparrow\)} & \textbf{LPIPS\(\downarrow\)} & \textbf{VBench\(\uparrow\)} \\
\midrule
Vanilla (50 steps) & 178.38 & 1.000\(\times\) & 24.53 & -- & -- & -- & 82.65 \\
Vanilla (25 steps) & 92.15 & 1.936\(\times\) & 24.53 & 0.6101 & 14.31 & 0.4151 & 83.01 \\
\midrule
TeaCache (\(\delta=0.25\)) & 68.58 & 2.601\(\times\) & 24.83 & 0.8574 & 23.98 & 0.1178 & 82.25 \\
MagCache (\(\delta=0.24,K=6\)) & 68.55 & 2.602\(\times\) & 24.83 & 0.8601 & 24.03 & 0.1153 & 82.39 \\
TaylorSeer (\(\mathcal{N}=7,O=2\)) & 67.84 & 2.629\(\times\) & 65.50 & 0.7932 & 21.15 & 0.1549 & 81.40 \\
HiCache (\(\mathcal{N}=6,O=2\)) & 68.80 & 2.593\(\times\) & 65.50 & 0.8316 & 23.57 & 0.1177 & 82.51 \\
VDE (\(n=5\)) & 68.36 & 2.609\(\times\) & \textbf{24.56} & 0.7698 & 20.93 & 0.1467 & 81.40 \\
\rowcolor{resultOurs}
\textbf{\method{} (Ours)} & \textbf{66.94} & \textbf{2.665\(\times\)} & \textbf{24.56} & \textbf{0.8793} & \textbf{25.16} & \textbf{0.1098} & \textbf{82.64} \\
\bottomrule
\end{tabularx}
\end{table*}

\Cref{tab:hunyuan-main,tab:wan-main} evaluate 65-frame
clips for the same 128 VBench prompts.
More visual results are provided in the supplement.

\paragraph{HunyuanVideo.}
\method{} delivers the highest fidelity on all three metrics
(\cref{tab:hunyuan-main}).
At \(2.71\times\), it is at least as fast as MagCache and VDE (\(2.63\times\))
and improves on MagCache, the strongest baseline, in SSIM (0.901 to 0.914)
and LPIPS (0.087 to 0.069).
TeaCache at its official threshold is slower (\(2.16\times\)) and still departs from the
full model, with 2.3 times the LPIPS and 5.2\,dB lower PSNR.
Second-order TaylorSeer and HiCache exceed the GPU memory.
Their first-order variants fit at matched speed (\(2.71\)--\(2.74\times\))
but trail \method{} by more than 3\,dB PSNR with 1.6--1.7 times its LPIPS,
while using 10\,GiB more memory.

\paragraph{Wan2.1.}
At a matched speed (\(2.66\times\) versus \(2.59\)--\(2.63\times\)), \method{}
attains the best SSIM, PSNR and LPIPS (\cref{tab:wan-main}).
Relative to MagCache and TeaCache at \(2.60\times\), it improves PSNR by
1.1--1.2\,dB and SSIM by about 0.02, and TaylorSeer and VDE trail it by
more than 4\,dB PSNR.

\begin{figure*}[!t]
\centering
\includegraphics[width=\textwidth]{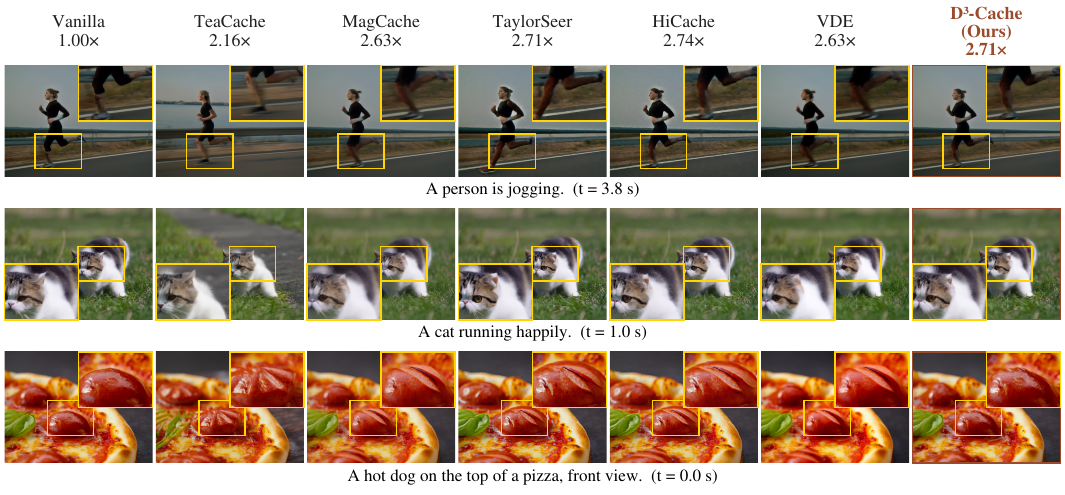}
\caption{\textbf{Qualitative comparison with previous acceleration methods on HunyuanVideo.}
One frame per clip, with the highlighted region magnified.
Compared methods blur or redraw the jogger's legs, the cat's face and the sausage, whereas
\method{} stays closest to the 50-step output.
TaylorSeer and HiCache use their first-order settings, as the second-order ones exceed the GPU memory.
Examples are taken from the prompts with the largest LPIPS advantage of \method{}.}
\label{fig:revision-hunyuan-qualitative}
\end{figure*}

\paragraph{Perceptual quality.}
The reference-free scores of \method{} stay close to those of the 50-step
model on all four models: ImageReward and CLIP score each drop by at most 0.03
on the image models, and the VBench total drops by 0.01 on Wan and 0.58 on
HunyuanVideo.
On Wan, \method{} attains the highest VBench total among compared cache-based methods
(82.64, versus 82.25--82.51 for TeaCache, MagCache and HiCache and 81.40 for
TaylorSeer and VDE); the 25-step model scores 83.01
despite its low fidelity, so VBench does not measure faithfulness to the full model.
On HunyuanVideo, it scores 82.06, above first-order HiCache (82.03), MagCache (81.84),
TeaCache (81.64) and VDE (80.98); first-order TaylorSeer scores highest (82.52)
despite 1.6 times the LPIPS.

\paragraph{Memory.}
\method{} adds negligible memory over the base model on all
four models, whereas second-order TaylorSeer and HiCache add 6.0--41.0\,GiB.

\paragraph{Fewer steps versus forecasting.}
Halving the number of steps loses far more fidelity: at about
\(2\times\), the 25-step models reach LPIPS 0.205--0.415, versus
0.069--0.218 for \method{} at \(2.6\)--\(3.8\times\).

\subsection{Analysis}\label{sec:results:controls}
\paragraph{Predictor under a common schedule.}
To isolate the predictor, we fix 16 full evaluations at
predictor-independent positions and compare predictors in the same velocity
space on 50 DrawBench prompts.
On HiDream, DMD with \(H=3\) lowers LPIPS to 0.245, from 0.299 for direct
reuse and 0.625 for two-point linear extrapolation, and raises PSNR by
1.1 and 7.6\,dB; both improvements are statistically significant.
On FLUX, it improves on linear extrapolation (LPIPS 0.234 versus 0.261,
PSNR \(+1.7\)\,dB) and is statistically indistinguishable from direct
reuse (0.241).
DMD with \(H=2\), \ie, global scaling, has higher LPIPS on both models
(0.264 on HiDream and 0.242 on FLUX); the difference is significant only on HiDream.
In the full system on HunyuanVideo (same gate and \(S\), all 128 prompts),
replacing DMD with direct reuse keeps 17.0 full evaluations per clip but
raises LPIPS from 0.069 to 0.104 and lowers
PSNR by 0.7\,dB; the SSIM, PSNR and LPIPS differences are significant.

\paragraph{History length.}
On HunyuanVideo, increasing \(H\) from 2 to 3 lowers LPIPS by 19--54\%
in seven tested \((S,\tau)\) settings at similar full-evaluation counts
(\eg, 0.121 to 0.075 with 17 in both).
Counts and other metrics do not improve uniformly; the supplement reports
all tested settings and the exceptions.

\paragraph{Gate under a fixed predictor.}
With the same \(H=3\) predictor and \(\tau=0.56\) on 50 DrawBench prompts,
the gate improves on fixed refresh intervals at similar cost.
On FLUX, it lowers LPIPS from 0.311 to 0.252 and raises PSNR by 2.1\,dB
with 12.2 versus 12 full evaluations, and from 0.273 to 0.233
(\(+1.7\)\,dB) with 13.4 versus 14.
On HiDream (\(S=3\)), LPIPS falls from 0.213 to 0.107 and PSNR rises by 5.8\,dB
with 20.1 versus 20.
All of these improvements are statistically significant.
How often the risk test, rather than the cap, ends a prediction run
depends on the model and on \(S\); the supplement reports both counts.

\paragraph{Qualitative comparison.}
\Cref{fig:revision-image-qualitative,fig:revision-hunyuan-qualitative}
compare outputs at the operating points of the main tables.
On HiDream, TaylorSeer and HiCache replace the forest behind the bear with
an orange backdrop, four of the five baselines lose the beer glass in the
stained-glass window, and TaylorSeer, HiCache and VDE flatten the cut
surface of the pears into a uniform yellow.
\method{} keeps the layout, colors and fine textures of the 50-step image.
On HunyuanVideo, TeaCache changes the scene altogether, and the other
baselines blur the jogger's legs, redraw the cat's face or alter the
sausage, whereas \method{} preserves both structure and detail.

\section{Conclusion}\label{sec:conclusion}

We presented \method{}, a training-free accelerator for Diffusion
Transformers that forecasts the velocity consumed by the scheduler with
Dynamic Mode Decomposition and refreshes through a closed-loop,
horizon-aware gate.
Forecasting only the velocity keeps the memory overhead
negligible regardless of network depth, and scoring every forecast against
the next exact velocity removes the need for a per-model offline calibration
curve: one history length and one risk tolerance serve four image and
video models, accelerated by \(2.6\)--\(3.8\times\).
At comparable speed, \method{} reproduces the base model output more
faithfully than the compared caching methods, most clearly on HiDream-I1 and HunyuanVideo,
whereas second-order activation forecasters need 6--41\,GiB more memory.

\bibliographystyle{plainnat}
\bibliography{references}

\end{document}